\documentclass[runningheads]{llncs}
\usepackage{graphicx}
\usepackage{tikz}
\usepackage{comment}
\usepackage{amsmath,amssymb} % define this before the line numbering.
\usepackage{color}

\usepackage[accsupp]{axessibility}  % Improves PDF readability for those with disabilities.

\usepackage{booktabs}
\usepackage{multirow}
\usepackage{makecell}
\usepackage{wrapfig}
\usepackage[misc]{ifsym}
\usepackage[pagebackref=false,breaklinks,colorlinks=true]{hyperref}

\begin{document}
% \linenumbers
\pagestyle{headings}
\mainmatter
\def\ECCVSubNumber{193}  % Insert your submission number here

\title{Graph R-CNN: Towards Accurate 3D Object Detection with Semantic-Decorated Local Graph} % Replace with your title

\begin{comment}
\titlerunning{ECCV-22 submission ID \ECCVSubNumber} 
\authorrunning{ECCV-22 submission ID \ECCVSubNumber} 
\author{Anonymous ECCV submission}
\institute{Paper ID \ECCVSubNumber}
\end{comment}
%******************

% \begin{comment}
\titlerunning{Graph R-CNN}
\author{Honghui Yang\inst{1}\index{Yang, Honghui}\and
Zili Liu\inst{1,3}\index{Liu, Zili}\and
Xiaopei Wu\inst{1}\index{Wu, Xiaopei}\and
Wenxiao Wang\inst{2}$^{(\textrm{\Letter})}$\index{Wang, Wenxiao}\and\\
Wei Qian\inst{3}\index{Qian, Wei}\and
Xiaofei He\inst{1,3}\index{He, Xiaofei}\and
Deng Cai\inst{1}\index{Cai, Deng}} 
\authorrunning{H. Yang et al.}
\institute{State Key Lab of CAD\&CG, Zhejiang University  \and
School of Software Technology, Zhejiang University  \and
Fabu Inc. \\
\email{\{yanghonghui,wuxiaopei,wenxiaowang,dcai\}@zju.edu.cn} \\
\email{\{zililiuzju\}@gmail.com} \\
\email{\{qianwei,hexiaofei\}@fabu.ai}
}
% \end{comment}
%******************
\maketitle

\begin{abstract}
	Two-stage detectors have gained much popularity in 3D object detection.
	Most two-stage 3D detectors utilize grid points, voxel grids, or sampled keypoints for RoI feature extraction in the second stage.
	Such methods, however, are inefficient in handling unevenly distributed and sparse outdoor points.
	This paper solves this problem in three aspects.
	1) Dynamic Point Aggregation.
	We propose the patch search to quickly search points in a local region for each 3D proposal.
	The dynamic farthest voxel sampling is then applied to evenly sample the points.
	Especially, the voxel size varies along the distance to accommodate the uneven distribution of points.
	2) RoI-graph Pooling.
	We build local graphs on the sampled points to better model contextual information and mine point relations through iterative message passing.
	3) Visual Features Augmentation.
	We introduce a simple yet effective fusion strategy to compensate for sparse LiDAR points with limited semantic cues.
	Based on these modules, we construct our Graph R-CNN as the second stage, which can be applied to existing one-stage detectors to consistently improve the detection performance.
	Extensive experiments show that Graph R-CNN outperforms the state-of-the-art 3D detection models by a large margin on both the KITTI and Waymo Open Dataset.
	And we rank the \textbf{1$^{st}$} place on the KITTI BEV car detection leaderboard.
	Code will be available at \url{https://github.com/Nightmare-n/GraphRCNN}.
	\end{abstract}

\section{Introduction}
In autonomous driving, 3D object detection is an essential task that has received substantial attention from industry~\cite{alex2020rcd,lang2019pointpillar,yan2018second} and academia~\cite{shi2020pvrcnn,yang20203dssd,zheng2021sessd}. Among the existing 3D detection methods, two-stage detectors~\cite{shi2019pointrcnn,yin2021center} outperform most single-stage detectors~\cite{lang2019pointpillar,yan2018second} in accuracy due to the proposal refinement stage. Previous two-stage methods~\cite{deng2021voxelrcnn,li2021lidarrcnn,mao2021pyramid,shi2020pvrcnn,shi2019pointrcnn,shi2020part,yang2019std} have explored different RoI pooling methods to capture better proposal features for refinement. PointRCNN~\cite{shi2019pointrcnn} and its subsequent works~\cite{li2021lidarrcnn} sample keypoints from original point clouds near the 3D proposal and extract the features of the sampled points. Part-A$^2$ Net~\cite{shi2020part} divides each 3D proposal into regular voxel grids and applies sparse convolutions to capture the features of voxel grids. PV-RCNN~\cite{shi2020pvrcnn} and its variants~\cite{deng2021voxelrcnn,mao2021pyramid,wu2022sfd} sample grid points within each 3D proposal and use the set abstraction~\cite{qi2017pointnet++} to aggregate the features of grid points. The methods that utilize the sampled keypoints show more flexibility than others since they directly process raw points and avoid the predefined voxel grids~\cite{liu2021sparsepoint,yang2019std} or grid points~\cite{cheng2021backtracing,wang2020infofocus} as intermediaries for RoI feature extraction.

\begin{figure}[!t]
	\centering
	\includegraphics[width=1.0\columnwidth]{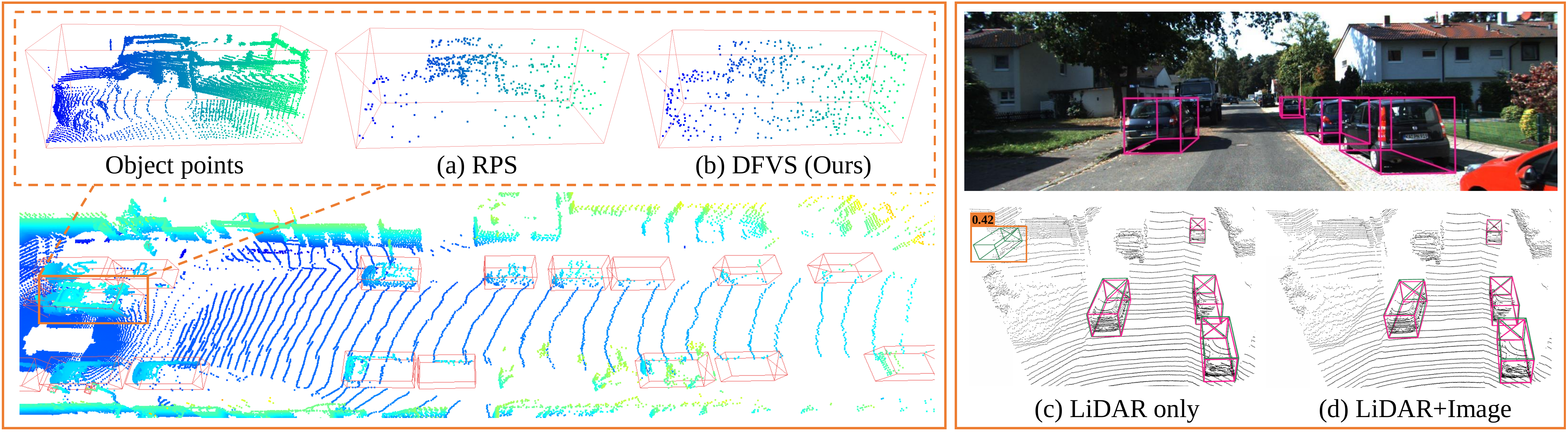}
	\caption{Illustration of different sampling strategies: (a) random point sampling (RPS) in existing works and the proposed (b) dynamic farthest voxel sampling (DFVS). And the comparison of results using (c) LiDAR and  (d) LiDAR and Image. We show the ground truth in pink bounding boxes and our detected objects in green bounding boxes.}
	\label{moti}
\end{figure}

Nevertheless, existing methods relied on sampled points still have some problems: \textbf{1)} They ignore that points are unevenly distributed in different parts of an object, thus yielding a suboptimal sampling strategy. As Fig.~\ref{moti}(a) shows, points for some parts are too sparse to preserve the structure information, which will hinder the prediction of the object's size. \textbf{2)} The point interrelation is not adequately utilized to model the contextual information of sparse points for object detection. \textbf{3)} Sparse LiDAR points in a single proposal provide limited semantic cues, which easily leads to a series of points that resemble a part of an object yielding high classification scores. Fig.~\ref{moti}(c) shows that a wall corner is wrongly detected as a car. 
To surmount the above challenges, we introduce three modules:

\textbf{1)} Dynamic Point Aggregation (DPA).
To efficiently and effectively group and sample context and object points for 3D proposals, we propose patch search (PS) and dynamic farthest voxel sampling (DFVS).
We will start with PS to speed up grouping then move on to DFVS to solve the uneven distribution problem during sampling. 

Previous methods~\cite{chen2019fastpointrcnn,shi2019pointrcnn,shi2020part} group points by searching all points to determine whether they belong to a proposal, which is time-consuming since the theoretical time complexity is $O(NM)$, where $N$ and $M$ are the number of points and proposals, respectively.
Especially for the detection on Waymo Open Dataset~\cite{sun2020wod}, there are often about 180K points and 500 proposals per frame that need to be processed.
In contrast, PS only searches the points falling in patches occupied by the proposal to group corresponding context and object points.
Then, DFVS sample keypoints from the grouped points to well retain the objects' structure, as shown in Fig.~\ref{moti}(b).
Specifically, instead of sampling points directly~\cite{li2021lidarrcnn,sheng2021ct3d,shi2019pointrcnn}, DFVS splits each proposal into evenly distributed voxels and iteratively samples the most distant non-empty voxels (i.e., voxels involving at least one point).
Further, to ensure the efficiency and accuracy of sampling, we resort to dynamic voxel size.
That is, for nearby objects with many points, we use a large voxel size to reduce the sampling complexity.
While for distant objects with sparse points, we use a relatively small voxel size to preserve the geometric details.

\textbf{2)} RoI-graph Pooling (RGP). % Point-GNN~\cite{shi2020pointgnn} demonstrates the ability of graph neural network (GNN) that could capture robust features from points for highly accurate object detection. 
To alleviate the missing detection, we utilize the graph neural network (GNN) to build connections among points to better model contextual information through iterative message passing. Compared with previous point-based methods~\cite{qi2017pointnet,qi2017pointnet++}, the GNN allows more complex features to be determined along the edges and avoids grouping and sampling the points repeatedly. Specifically, RGP constructs local graphs in each 3D proposal, which treats the sampled points as graph nodes. To compensate for the information loss caused by downsampling, we use PointNet~\cite{qi2017pointnet} to encode neighbor points of each node into the initial features. Then, RGP iteratively aggregates messages from its neighbors on a k-NN graph to mine the relations among nodes. Finally, we propose multi-level attentive fusion (MLAF) to capture abundant spatial features from multi-level nodes with different receptive fields and fully exploit graph nodes to extract robust RoI features.

\textbf{3)} Visual Features Augmentation (VFA). Though LiDAR points provide accurate depth information, the lack of sufficient semantic features makes it difficult to distinguish objects with similar geometric structures. Thus, a simple yet effective fusion method is used to fuse geometric features from LiDAR and semantic features from images for suppressing the false positives, as shown in Fig.~\ref{moti}(d). We decorate local graphs with image features by bilinear interpolation since graph nodes serve as a natural bridge between the LiDAR and image. We train the two streams in an end-to-end manner and show the complex multi-modality cut-and-paste augmentation~\cite{wang2021pointaug,zhang2020moca} is not necessary for our framework.

Based on the three modules, we present our Graph R-CNN that can replace the second stage of other two-stage detectors or supplement any one-stage detector for further improvement.
Extensive experiments have been conducted on several detection benchmarks to verify the effectiveness of our approach.
We consistently improve existing 3D detectors by a large margin and achieve new state-of-the-art results on both Waymo Open Dataset (WOD)~\cite{sun2020wod} and KITTI~\cite{geiger2012kitti}.

Our contributions can be summarized as follows:
\begin{itemize}
	\item We fully consider the uneven distribution of point clouds and propose dynamic point aggregation (DPA).
	\item We introduce RoI-graph Pooling (RGP) to capture the robust RoI features by iterative graph-based message passing.
	\item We demonstrate a simple yet effective fusion strategy (VFA) to fuse image features with point features during the refinement stage.
	\item We present an accurate and efficient 3D object detector (Graph R-CNN) that can be applied to existing 3D detectors. Extensive experiments are conducted to verify the effectiveness of our methods.
\end{itemize}

\section{Related Works}
\paragraph{3D Object Detection Using Point Cloud.} % The performance of 3D object detectors continues to make breakthroughs with the development of depth sensors and deep learning. %Due to the difficulty of multi-sensor fusion, mainstream 3D object detectors directly detect 3D objects on 3D point clouds and have made great progress. 
Current 3D detectors can be mainly divided into two streams: one-stage and two-stage methods. One-stage detectors jointly predict an output class and location of objects at the projected volumetric grids or downsampled points. % VoxelNet~\cite{zhou2018voxelnet} rasterizes point cloud into 3D voxels and leverages PointNet~\cite{qi2017pointnet} to generate a voxel-wise representation, followed by 3D CNNs and 2D CNNs for objects detection. 
SECOND~\cite{yan2018second} rasterizes point cloud into 3D voxels and accelerates VoxelNet~\cite{zhou2018voxelnet} by exploiting sparse 3D convolution. PointPillars~\cite{lang2019pointpillar} partitions points into pillars rather than voxels. 3DSSD~\cite{yang20203dssd} proposes a fusion farthest point sampling strategy by utilizing both feature and geometry distance for better classification performance. 

Two-stage detectors first use a region proposal network (RPN) to generate coarse object proposals and then use a dedicated per-region head to classify and refine them. PointRCNN~\cite{shi2019pointrcnn} generates RoIs based on foreground points from the scene and conducts canonical 3D box refinement after point cloud region pooling. PV-RCNN~\cite{shi2020pvrcnn} incorporates the advantage from 3D voxel Convolutional Neural Network and Point-based set abstraction to learn discriminative point cloud features. Voxel R-CNN~\cite{deng2021voxelrcnn} proposes a voxel RoI pooling to extract RoI features directly from voxel features to refine proposals in the second stage. % CT3D~\cite{sheng2021ct3d} proposes a second stage that extracts features based on raw point clouds and prediction boxes, making it not dependent on a specific one-stage method. 
CenterPoint~\cite{yin2021center} detects centers of objects using a keypoint detector and refines these estimates using additional point features on the object.

\paragraph{3D Object Detection Using Multi-modality Fusion.}
Recently, much progress has been made to exploit the advantages of the camera and LiDAR sensors. MV3D~\cite{chen2017mv3d} generates 3D proposals from the bird’s eye view and fuses multi-view features via region-based representation. EPNet~\cite{huang2020epnet} proposes LI-Fusion module to fuse the deep features of point
clouds and camera images in a point-wise paradigm. However, insufficient multi-modality augmentation makes these methods perform only marginally better or sometimes worse than approaches that only use point cloud. Recent works~\cite{wang2021pointaug,zhang2020moca} overcome the constraint by extending the cut-and-paste augmentation~\cite{yan2018second} to multi-modality methods. But a complex process is needed to avoid collisions between objects in both point cloud and 2D imagery domain. PointPainting~\cite{vora2020pointpainting} augments LiDAR points with segmentation scores, which is suboptimal to cover color and textures in images.

\paragraph{3D Object Detection Using Graph Neural Networks.} Graph Neural Networks~\cite{m2005gnn} are introduced to model intrinsic relationships of graph-structured data. Since they are suitable for processing 3D point clouds, some works have adopted GNNs for 3D object detection. 3DVID~\cite{yin20203dvid} explores spatial relations among different grid regions by treating the non-empty pillar grids as graph nodes to enhance pillar features. Object DGCNN~\cite{wang2021objectdgcnn} uses DGCNN~\cite{wang2019dgcnn} to construct a graph between the queries for incorporating neighborhood information in object detection estimates. DOPS~\cite{najibi2020dops} creates a graph where the points are connected to those with similar center predictions for consolidating the per-point object predictions. Point-GNN~\cite{shi2020pointgnn} encodes point clouds in a fixed radius near-neighbors graph and predicts the category and shape of the object that each node in the graph belongs to. Our work differs from previous works by constructing local graphs during the refinement stage, which greatly saves computational and memory overhead since the k-nearest neighbor algorithm can be parallelly applied in each 3D proposal, and numerous background points can be avoided to build graphs.

\section{Methods}
\begin{figure*}[!t]
	\centering
	\includegraphics[width=1\columnwidth]{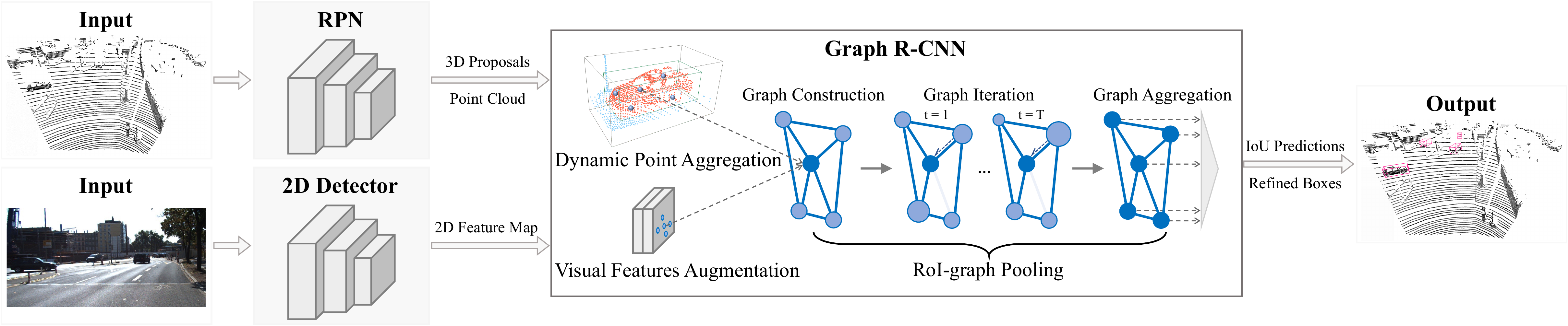}
	\caption{The overall architecture. We take 3D proposals and points from the region proposal network (RPN) and 2D feature map from the 2D detector as inputs. We propose dynamic point aggregation to sample context and object points and visual features augmentation to decorate the points with 2D features. RoI-graph pooling serves sampled points as graph nodes to build local graphs for each 3D proposal. We iterate the graph for $T$ times to mine the geometric features among the nodes. Finally, each node is fully utilized through graph aggregation to produce robust RoI features.}
	\label{stru}
\end{figure*}

In this section, we present the design of Graph R-CNN, as shown in Fig.~\ref{stru}. We first introduce dynamic point aggregation in Sec.~\ref{subsection:dpa}. Next, we will demonstrate RoI-graph pooling in Sec.~\ref{subsection:rgp}. Then, we will illustrate how to incorporate semantic features from the image into our framework in Sec.~\ref{subsection:vfa}. Finally, we will show the definition of the loss function in Sec.~\ref{subsection:loss}.

\subsection{Dynamic Point Aggregation} \label{subsection:dpa}
\begin{figure}[!t]
	\centering
	\includegraphics[width=0.95\columnwidth]{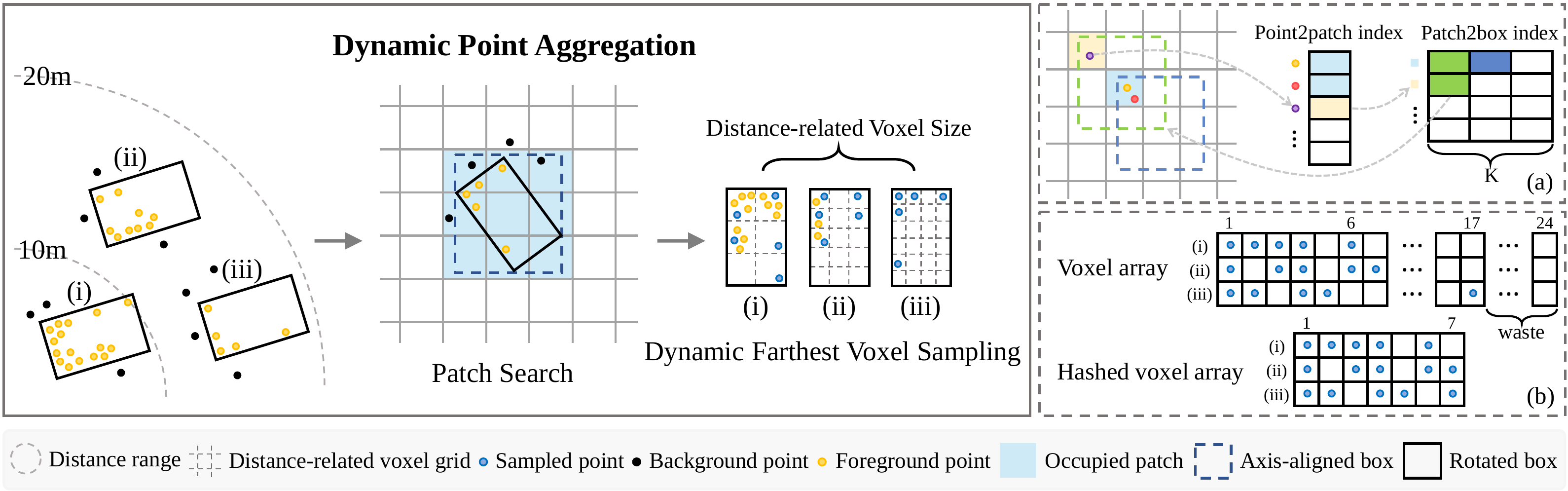}
	\caption{Illustration of dynamic point aggregation, which includes (a) patch search and (b) dynamic farthest voxel sampling. In (a), we use different colors to represent different keys and values. In (b), we flatten the voxel grids of each proposal for better display.}
	\label{dpa}
\end{figure}

In this section, we present a differentiable dynamic point aggregation (DPA) to efficiently and effectively group and sample points and their features for each proposal.
We first enlarge each proposal's size by $\sigma$ to wrap enough object and context points.
Then, DPA uses patch search (PS) to quickly group the points in each enlarged proposal and dynamic farthest voxel sampling (DFVS) to evenly sample the grouped points.
\paragraph{Patch Search.}
Unlike previous works~\cite{li2021lidarrcnn,sheng2021ct3d,shi2019pointrcnn} that need to search all points to determine whether they belong to an enlarged proposal, we divide the entire scene into patches and only search the points falling in patches occupied by the proposal, as shown in Fig.~\ref{dpa}.
PS consists of three major steps: 1) We turn the rotated box into an axis-aligned box to make it easier to find the occupied patches.
2) We build \texttt{point2patch} and \texttt{patch2box} index arrays, which store the point and patch indices as keys, and the corresponding patch and box indices as values, respectively.
3) We finally group the points for each proposal according to the \texttt{point2patch} and \texttt{patch2box} index arrays, as shown in Fig.~\ref{dpa}(a).
We note that all the steps can be conducted in parallel on GPUs.
In this way,  we reduce the theoretical time complexity from $O(NM)$ to $O(QK)$, where $Q$ is the number of points that fall in all occupied patches, and $K$ is the predefined maximum number of boxes per patch since the same patch may be occupied by multiple boxes. Notably, $Q$ and $K$ are much smaller than $N$ and $M$, respectively.

\begin{figure}[!t]
	\centering
	\includegraphics[width=0.8\columnwidth]{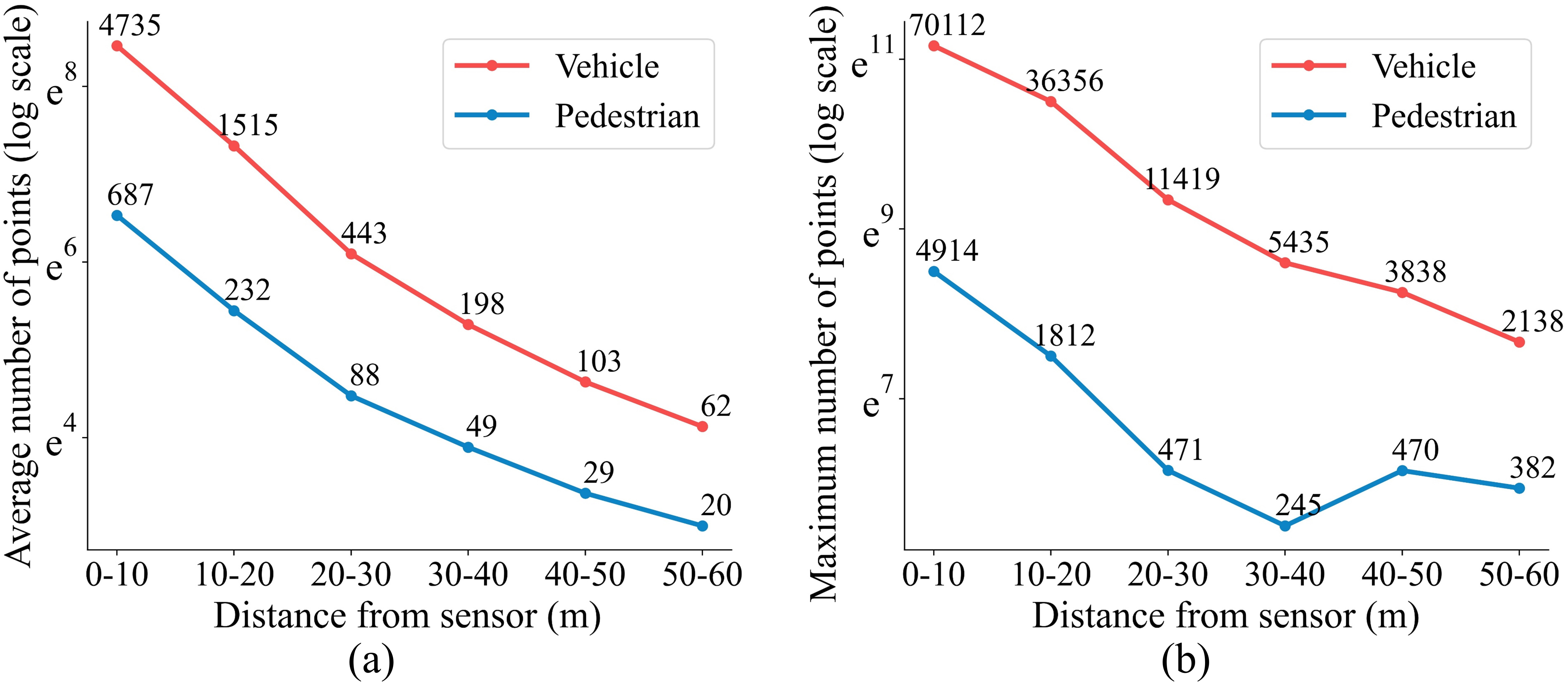}
	\caption{The statistical plot of the (a) average and (b) maximum number of points in each ground truth on Waymo Open Dataset for vehicle and pedestrian.}
	\label{statis}
\end{figure}

\paragraph{Dynamic Farthest Voxel Sampling.}
Since the number of raw points in a box is usually far more than that of the sampled points (e.g., 70112 vs. 256, as Fig.~\ref{statis}(b) shows), it's nontrivial to ensure every part of an object is sampled.
Therefore, we propose DFVS to balance sampling efficiency and accuracy.
To be specific, DFVS partitions proposals into evenly distributed voxels and then iteratively sample the most distant non-empty voxels.
Considering the number of points varies with the distance of the box, as Fig.~\ref{statis}(a) shows, the voxel size should be changed dynamically according to the distance to ensure the sampling efficiency of nearby objects and accuracy of distant objects, as shown in Fig.~\ref{dpa}. Formally, the voxel size $V_i$ of the box $b_i$ can be calculated by:
\begin{equation}
V_i = \lambda \cdot e^{-\frac{\sqrt{x_i^2 + y_i^2 + z_i^2}}{\delta}}
\end{equation}
where $(x_i, y_i, z_i)$ is the $i$-th box's center, and $\lambda$ and $\delta$ determine the relationship between the voxel size and the distance from the box to the LiDAR sensor. 

Assume we have grouped the points in the box $b_i$ by patch search and obtained $\mathcal{P}_i=\{p_j^i=\left[x_j^i, y_j^i, z_j^i, r_j^i\right] \in \mathbb{R}^4: j = 1,\cdots,N\}$, where $(x_j^i, y_j^i, z_j^i)$ and $r_j^i$ indicate $j$-th point's coordinate in the $i$-th box's canonical coordinate system and the reflectance intensity.
We assign each point to evenly divided voxel grids, and the grid index of the point $p_j^i$ is represented as $\{g_j^i=(\lfloor\frac{x_j^i}{V_i}\rfloor, \lfloor\frac{y_j^i}{V_i}\rfloor, \lfloor\frac{z_j^i}{V_i}\rfloor)\}$.
Each non-empty voxel can be represented by a randomly selected point in the voxel. Next, farthest point sampling (FPS)~\cite{qi2017pointnet++} is applied to iteratively sample the most distant non-empty voxels.

A potential problem with DVFS lies in that, for the distant box, a small voxel size will divide the box into numerous voxels, of which the non-empty voxels only occupy a small part. And to utilize the parallel computation of GPUs, voxel grids of other boxes need to be padded to the largest grid number, which will increase the memory overhead, as shown in Fig.~\ref{dpa}(b). Since we only care about non-empty voxels, we use a hash table~\cite{mao2021votr} to record the hashed grid indices of non-empty voxels and quadratic probing to resolve the collisions in the hash table.

\subsection{RoI-graph Pooling} \label{subsection:rgp}
In this section, we describe the process of RoI-graph pooling, as shown in Fig.~\ref{stru}, which treats sampled points as nodes to build local graphs in 3D proposals. It consists of graph construction, iteration, and aggregation.

\paragraph{Graph Construction.}
Given sampled points $\overline{\mathcal{P}}=\{p_j=\left[x_j, y_j, z_j, r_j\right] \in \mathbb{R}^4: j = 1,\cdots,T\}$ for each proposal $b$ (we drop the $i$ subscript for ease of notation), we construct a local graph $\mathcal{G} = (\mathcal{V}, \mathcal{E})$, where node $v_j\in \mathcal{V}$ represents a sampled point $p_j\in\overline{\mathcal{P}}$, and edge $e_j^k\in \mathcal{E}$ indicates the connection between node $v_j$ and $v_k$. To reduce the computational overhead, we define $\mathcal{G}$ as a k-nearest neighbor (k-NN) graph, which is built from the geometric distance among different nodes. Despite efficient, building graphs on down-sampled points inevitably loss fine-grained features. Thus, we use PointNet~\cite{qi2017pointnet} to encode original neighbor points within a radiu $r$ for each node. We note that neighbor query only induces a marginal computational overhead 
because it is only conducted for each proposal. 

The same graph nodes may be wrapped by different proposals, which will result in the same pooling features and thus introduce ambiguity in the refinement stage~\cite{li2021lidarrcnn,shi2020part}. Inspired by~\cite{sheng2021ct3d}, we add the 3D proposal's local corners (i.e., the corners are transformed to the proposal's canonical coordinate system) for each node to make them have the ability to discriminate differences. In our experiments, we found that two diagonal corners are sufficient. Formally, the initial state $s_j^0$ of each node $v_j$ at iteration step $t=0$ can be represented by:
\begin{equation}
s_j^0 = \left[x_j, y_j, z_j, r_j, f_j, u_j, w_j\right],
\end{equation}
where $\left[\cdot,\cdot\right]$ is concatenation function, $f_j$ is the features from PointNet, and $u_j$ and $w_j$ are two diagonal corners of the 3D proposal.  

\begin{figure}[!t]
	\centering
	\includegraphics[width=0.75\columnwidth]{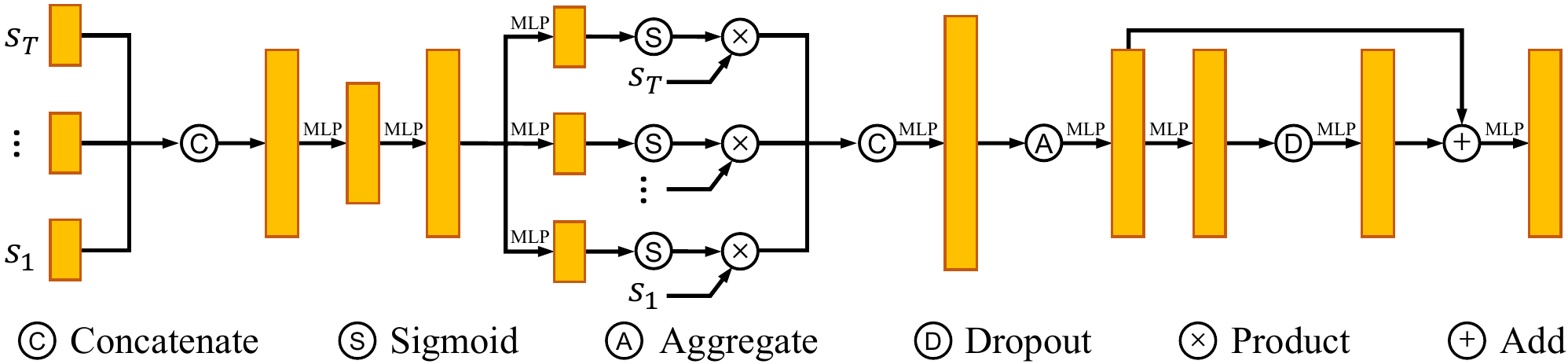}
	\caption{Illustration of multi-level attentive fusion.}
	\label{mlaf}
\end{figure}

\paragraph{Graph Iteration.}
To mine the rich geometric relations among nodes, we iteratively pass the message on $\mathcal{G}$ and update the node's state at each iteration step. Concretely, at step $t$, a node $v_j$ aggregates information from all the neighbor nodes $v_k \in \mathcal{N}_{v_j}$  in the k-NN graph. Following~\cite{chai2021ppc,wang2021objectdgcnn,yin20203dvid}, we use EdgeConv~\cite{wang2019dgcnn} to update the state $s_j^{t+1}$:

\begin{equation}
s_j^{t+1} = \max_{v_k \in \mathcal{N}_{v_j}}{\phi_\theta([s_k^t-s_j^t, s_j^t])},
\end{equation}
where $\phi_\theta$ is parameterized by a Multilayer Perceptron (MLP).

\paragraph{Graph Aggregation.}
To capture robust RoI features, we propose multi-level attentive fusion (MLAF) to fuse the nodes' features, as shown in Fig.~\ref{mlaf}. Specifically, we concatenate the nodes' features $[s_j^1, \cdots, s_j^T]$ from different iterations and feed them into several MLPs to learn the channel-wise weights. Then, we reweight $[s_j^1, \cdots, s_j^T]$ to enhance the features for final detection. After that, it's nontrivial to fully utilize every graph node for the proposal refinement. We explore several aggregation operations, e.g., the channel-wise Transformer~\cite{sheng2021ct3d}, Set Transformer~\cite{lee2019settrans}, attention sum, average pooling, and max pooling. Our final model uses max pooling as it provides the best empirical performance. Then, Dropout is used in later MLPs to reduce overfitting, and a shortcut connection is added to fuse more features without adding much cost.

\subsection{Visual Features Augmentation} \label{subsection:vfa}
Cut-and-paste augmentation (CPA)~\cite{yan2018second} is widely used for 3D object detection to increase the training samples, which could speed up training convergence and improve the detection performance.
Since our Graph R-CNN extracts RoI features directly from raw point clouds, we doubt whether CPA is required for our model.
We carefully study its influence in Sec.~\ref{subsection:as} and find that our model does not depend on it.
Thus, CPA is only used to pretrain the RPN and disabled when training the whole framework.

For the camera image, we extract high-level semantic features using a pretrained 2D detector.
Then, we apply two $1\times1$ convolutional kernels to reduce the dimensionality of the output feature, as Fig.~\ref{stru} shows.
The benefit brought by it is twofold. Firstly, it can learn to select features that contribute greatly to the performance of the refinement.
Secondly, it can ease the optimization to fuse low-dimensionality point features with high-dimensionality image features.
Then, we project the graph node to the location in the camera image and collect the feature vector at that pixel in the camera image through bilinear interpolation.
Lastly, the features will be appended to $s_j^0$ for each node $v_j$.

\subsection{Loss Functions} \label{subsection:loss}
\paragraph{Classification Loss.} For class-agnostic confidence score prediction, we follow~\cite{shi2020pvrcnn,yin2021center} to use a score target $I_i$ guided by the box’s 3D IoU with the corresponding ground-truth bounding box:
\begin{equation}
I_i = \min \left(1, \max \left(0, 2\times \mathrm{IoU}_{i} - 0.5\right) \right),
\end{equation}
where $\mathrm{IoU}_i$ is the IoU between the $i$-th proposal box and the ground truth. The training is supervised with a binary cross entropy loss:
\begin{equation}
\mathcal{L}_{cls}=\frac{1}{B} \sum_{i=1}^{B} -I_{i} \log \left(\hat{I}_{i}\right)-\left(1-I_{i}\right) \log \left(1-\hat{I}_{i}\right),
\end{equation}
where $\hat I_i$ is the predicted confidence score, and $B$ is the number of sampled region proposals at the training stage.

\paragraph{Regression Loss.} For box prediction, we transform the 3D proposal $b_i = (x_i, y_i, z_i,\\ l_i, w_i, h_i, \theta_i)$ and the corresponding 3D ground-truth bounding box $b_i^{gt} = (x_i^{gt}, y_i^{gt},\\ z_i^{gt}, l_i^{gt}, w_i^{gt}, h_i^{gt}, \theta_i^{gt})$ from the global reference frame to the canonical coordinate system of 3D proposal:
\begin{equation}
\begin{aligned}
& \tilde{b}_{i}=\left(0,0,0, l_{i}, w_{i}, h_{i}, 0\right), \\
& \tilde{b}_{i}^{gt}=\left(x_{i}^{gt}-x_{i}, y_{i}^{gt}-y_{i}, z_{i}^{gt}-z_{i}, l_{i}^{gt}, w_{i}^{gt}, h_{i}^{gt}, \Delta \theta_{i}\right),
\end{aligned}
\end{equation}
where $\Delta \theta_{i}=\theta_{i}^{gt}-\theta_{i}$. Then, the regression targets for center $t_{i}^{c}$, size $t_{i}^{s}$, and orientation $t_{i}^{o}$ can be defined as:
\begin{equation}
\begin{aligned}
& t_{i}^{c}=\left(x_{i}^{gt}-x_{i}, y_{i}^{g t}-y_{i}, z_{i}^{g t}-z_{i}\right), \\
& t_{i}^{s}=\left(l_{i}^{gt}-l_{i}, w_{i}^{gt}-w_{i}, h_{i}^{gt}-h_{i}\right), \\
& t_{i}^{o}=\Delta \theta_{i} - \lfloor\frac{\Delta \theta_{i}}{\pi} + 0.5\rfloor \times \pi.
\end{aligned}
\end{equation}

Having all the targets $t_i=(t_i^c,t_i^s,t_i^o)$, our regression loss is defined as:
\begin{equation}
\mathcal{L}_{reg}=\frac{1}{B_{+}} \sum_{i=1}^{B_{+}} \operatorname{L1}\left(o_{i}-t_{i}\right),
\end{equation}
where $o_i$ is the output of the model’s regression branch, and $B_{+}$ is the number of positive samples.

\paragraph{Total Loss.} Finally, the overall loss is formulated as:
\begin{equation}
\mathcal{L}=\mathcal{L}_{cls}+\alpha \mathcal{L}_{reg},
\end{equation}
where $\alpha$ is a hyperparameter to balance the loss, which is $1$ by default.

\setcounter{footnote}{0}
\section{Experiments}
\subsection{Datasets}

\paragraph{Waymo Open Dataset.}
is a large-scale autonomous driving dataset consisting of 798 scenes for training and 202 scenes for validation.
The evaluation protocol consists of average precision (AP) and average precision weighted by heading (APH).
It includes two difficulty levels: LEVEL\_1 denotes objects containing more than 5 points, and LEVEL\_2 denotes objects containing at least 1 point.

\paragraph{KITTI.}
contains 7481 training samples and 7518 testing samples in autonomous driving scenes.
We follow~\cite{chen2017mv3d,deng2021voxelrcnn,shi2020pvrcnn} to divide the training data into a $train$ set with 3712 samples and a $val$ set with 3769 samples.
The performance on the $val$ set and the test leaderboard are reported. 

\subsection{Implementation Settings}
\paragraph{Implementation Details.}
The codebase of CenterPoint is used for WOD. Then, we replace the second stage of CenterPoint-Voxel with our method (i.e., Graph-Ce) and train the network separately.
For the dynamic point aggregation, we sample $256$ points for each proposal and set $\sigma=0.4$. In dynamic farthest voxel sampling, we set hash size as $4099$ and $\lambda$ and $\delta$ as $0.18$ and $50$, respectively. In patch search, the $K$ and patch size are set to $32$ and $1.0$, respectively. For the RoI-graph pooling, we set $r=0.4$ and the embedding channels to $[16, 16]$ in PointNet. We update the graph with $T=3$, and the output dimensions of the three iterations are $[32, 32, 64]$. The number $k$ of nearest neighbors is set as $8$. In MLAF, the embedding dimension is $256$, and the dropout ratio is $0.1$.

For KITTI, the codebase of OpenPCDet is used.
We propose Graph-Pi, Graph-Vo, and Graph-Po that use the \textbf{pi}llar-based PointPillars, the \textbf{vo}xel-based SECOND, and the \textbf{po}int-based 3DSSD as their region proposal networks, respectively.
We incorporate the image branch in Graph-Vo (i.e., Graph-VoI) to compare with previous multi-modality methods.
For 2D detector, we use the CenterNet~\cite{zhou2019centernet} with DLA-34~\cite{yu2018dla} backbone, which takes images with a resolution of $1280\times384$ as input.
The dimension of the output features will be reduced to $32$ by the feature reduction layer.

\paragraph{Training Details.}
For WOD, we use the same training schedules and assignment strategies as CenterPoint-Voxel. The second stage is trained for $6$ epochs on 4 GTX 1080Ti GPUs with $8$ batch size per GPU. 

For KITTI, we use the same training configuration as PV-RCNN and train the whole model end-to-end for $80$ epochs on 4 GTX 1080Ti GPUs with $4$ batch size per GPU, and the pretrained RPN and 2D detector are frozen during training. For 2D detector, we pretrain CenterNet on WOD for 24 epochs and finetune it on KITTI for 12 epochs. We use Adam optimizer with one-cycle policy and set batch size to $2$ and learning rate to $0.00025$.

\begin{table*}[tbp]
	\centering
	\caption{Vehicle detection results on WOD validation sequences. CenterPoint-Voxel$^\dag$ is reproduced by us based on the officially released code. CenterPoint-Voxel$^\ddag$ is the first stage of CenterPoint-Voxel$^\dag$.}
	\resizebox{1.0\columnwidth}!{\begin{tabular}{c|c||cccc|cccc|cccc|cccc}
			\hline
			\multirow{2}{*}{Difficulty} & \multirow{2}{*}{Methods} & \multicolumn{4}{c|}{3D AP (IoU=0.7)} & \multicolumn{4}{c|}{3D APH (IoU=0.7)} & \multicolumn{4}{c|}{BEV AP (IoU=0.7)} & \multicolumn{4}{c}{BEV APH (IoU=0.7)} \\
			&  & Overall & 0-30m & 30-50m & 50m-Inf & Overall & 0-30m & 30-50m & 50m-Inf & Overall & 0-30m & 30-50m & 50m-Inf & Overall & 0-30m & 30-50m & 50m-Inf \\
			\hline
			\multirow{10}{*}{LEVEL 1}
			& MVF~\cite{zhou2019mvf} & 62.93 & 86.30 & 60.02 & 36.02 & - & - & - & - & 80.40 & 93.59 & 79.21 & 63.09 & - & - & - & - \\
			& Pillar-od~\cite{wang2020piilar-od} & 69.80 & 88.53 & 66.50 & 42.93 & - & - & - & - & 87.11 & 95.78 & 84.74 & 72.12 & - & - & - & - \\
			& PV-RCNN~\cite{shi2020pvrcnn} & 70.30 & 91.92 & 69.21 & 42.17 & 69.69 & 91.34 & 68.53 & 41.31 & 82.96 & 97.35 & 82.99 & 64.97 & 82.06 & 96.71 & 82.01 & 63.15 \\ 
			& VoTr-TSD~\cite{mao2021votr} & 74.95 & 92.28 & 73.36 & 51.09 & 74.25 & 91.73 & 72.56 & 50.01 & - & - & - & - & - & - & - & - \\
			& Voxel R-CNN~\cite{deng2021voxelrcnn} & 75.59 & 92.49 & 74.09 & 53.15 & - & - & - & - & 88.19 & \bf{97.62} & 87.34 & 77.70 & - & - & - & - \\
			& LiDAR R-CNN~\cite{li2021lidarrcnn} & 76.00 & 92.10 & 74.60 & 54.50 & 75.50 & 91.60 & 74.10 & 53.40 & 90.10 & 97.00 & 89.50 & 78.90 & 89.30 & 96.50 & 88.60 & 77.40 \\
			& Pyramid-PV~\cite{mao2021pyramid} & 76.30 & 92.67 & 74.91 & 54.54 & 75.68 & 92.20 & 74.21 & 53.45 & - & - & - & - & - & - & - & - \\
			& CenterPoint-Voxel$^\dag$~\cite{yin2021center} & 76.86 & 92.27 & 75.31 & 54.10 & 76.33 & 91.81 & 74.74 & 53.35 & 91.61 & 97.19 & 91.05 & 82.06 & 90.85 & 96.69 & 90.23 & 80.59 \\
			\cline{2-18}
			& CenterPoint-Voxel$^\ddag$~\cite{yin2021center} & 74.78 & 91.51 & 73.25 & 50.67 & 74.24 & 91.04 & 72.67 & 49.93 & 90.94 & 97.02 & 90.26 & 80.78 & 90.12 & 96.51 & 89.39 & 79.20 \\
			& \textbf{Graph-Ce (Ours)} & \textbf{80.77} & \textbf{93.59} & \textbf{79.68} & \textbf{60.41} & \textbf{80.28} & \textbf{93.20} & \textbf{79.16} & \textbf{59.62} & \textbf{92.69} & 97.56 & \textbf{92.15} & \textbf{84.31} & \textbf{92.01} & \textbf{97.15} & \textbf{91.43} & \textbf{82.94} \\
			\hline
			\multirow{8}{*}{LEVEL 2}
			& PV-RCNN~\cite{shi2020pvrcnn} & 65.36 & 91.58 & 65.13 & 36.46 & 64.79 & 91.00 & 64.49 & 35.70 & 77.45 & 94.64 & 80.39 & 55.39 & 76.60 & 94.03 & 79.40 & 53.82 \\
			& VoTr-TSD~\cite{mao2021votr} & 65.91 & - & - & - & 65.29 & - & - & - & - & - & - & - & - & - & - & -  \\
			& Voxel R-CNN~\cite{deng2021voxelrcnn} & 66.59 & 91.74 & 67.89 & 40.80 & - & - & - & - & 81.07 & \textbf{96.99} & 81.37 & 63.26 & - & - & - & - \\
			& Pyramid-PV~\cite{mao2021pyramid} & 67.23 & - & - & - & 66.68 & - & - & - & - & - & - & - & - & - & - & - \\
			& LiDAR R-CNN~\cite{li2021lidarrcnn} & 68.30 & 91.30 & 68.50 & 42.40 & 67.90 & 90.90 & 68.00 & 41.80 & 81.70 & 94.30 & 82.30 & 65.80 & 81.00 & 93.90 & 81.50 & 64.50 \\
			& CenterPoint-Voxel$^\dag$~\cite{yin2021center} & 69.09 & 91.41 & 69.43 & 42.40 & 68.59 & 90.96 & 68.89 & 41.78 & 85.43 & 96.35 & 86.44 & 70.06 & 84.66 & 95.86 & 85.63 & 68.66 \\
			\cline{2-18}
			& CenterPoint-Voxel$^\ddag$~\cite{yin2021center} & 66.66 & 90.63 & 66.90 & 39.50 & 66.17 & 90.16 & 66.36 & 38.90 & 84.87 & 96.21 & 85.69 & 69.08 & 84.04& 95.69 & 84.81 & 67.58 \\
			& \textbf{Graph-Ce (Ours)} & \textbf{72.55} & \textbf{92.75} & \textbf{73.74} & \textbf{47.84} & \textbf{72.10} & \textbf{92.36} & \textbf{73.25} & \textbf{47.19} & \textbf{86.56} & 96.79 & \textbf{87.59} & \textbf{72.06} & \textbf{85.86} & \textbf{96.38} & \textbf{86.86} & \textbf{70.72} \\
			\hline
	\end{tabular}}
	\label{overall_results}
\end{table*}

\begin{table*}[tbp]
	\centering
	\caption{Vehicle, pedestrian, and cyclist results on WOD validation sequences.}
	% \vspace{-0.8em}
	\resizebox{1.0\columnwidth}!{\begin{tabular}{c|c||cccc|cccc|cccc}
			\hline
			\multirow{2}{*}{Difficulty} & \multirow{2}{*}{Methods} & \multicolumn{4}{c|}{Vehicle} & \multicolumn{4}{c|}{Pedestrian} & \multicolumn{4}{c}{Cyclist} \\
			&  & 3D AP & 3D APH & BEV AP & BEV APH & 3D AP & 3D APH & BEV AP & BEV APH & 3D AP & 3D APH & BEV AP & BEV APH \\
			\hline
			\multirow{2}{*}{LEVEL 1}
			& CenterPoint-Voxel$^\ddag$~\cite{yin2021center} & 74.78 & 74.24 & 90.94 & 90.12 & 75.95 & 69.75 & 82.01 & 75.05 & 72.27 & 71.12 & 75.95 & 74.70 \\
			& \textbf{Graph-Ce (Ours)} & \textbf{80.77} & \textbf{80.28} & \textbf{92.69} & \textbf{92.01} & \textbf{82.35} & \textbf{76.64} & \textbf{86.75} & \textbf{80.51} & \textbf{75.28} & \textbf{74.21} & \textbf{77.42} & \textbf{76.30} \\
			\hline
			\multirow{2}{*}{LEVEL 2}
			& CenterPoint-Voxel$^\ddag$~\cite{yin2021center} & 66.66 & 66.17 & 84.87 & 84.04 & 68.42 & 62.67 & 75.06 & 68.46 & 69.69 & 68.59 & 73.24 & 72.03 \\
			& \textbf{Graph-Ce (Ours)} & \textbf{72.55} & \textbf{72.10} & \textbf{86.56} & \textbf{85.86} & \textbf{74.44} & \textbf{69.02} & \textbf{79.50} & \textbf{73.45} & \textbf{72.52} & \textbf{71.49} & \textbf{74.64} & \textbf{73.56} \\
			\hline
	\end{tabular}}
	% \vspace{-1.5em}
	\label{all_class_results}
\end{table*}

\subsection{Comparison with State-of-the-Art Methods} \label{subsection:sota}
\paragraph{Waymo Open Dataset.} We compare Graph-Ce for the vehicle class at different distances on the full WOD validation with previous methods. Table~\ref{overall_results} shows that Graph-Ce achieves the state-of-the-art results in both level 1 and level 2 among all the published papers with a single frame LiDAR input. In Table~\ref{all_class_results}, we present our results for the vehicle, pedestrian, and cyclist classes. Compared with our baseline, i.e., CenterPoint-Voxel$^\ddag$, our method improves the 3D APH in level 2 for the vehicle, pedestrian, and cyclist by 5.93\%, 6.35\%, and 2.90\%, respectively.% Moreover, we visualize the detection results in Fig.~\ref{demo}.

\paragraph{KITTI.}
We compare Graph-Pi, Graph-Vo, Graph-Po, and Graph-VoI with previous methods listed in Table~\ref{kitti_test}.
Graph-Pi achieves the fastest inference speed among all two-stage methods.
Compared with methods using only LiDAR as input, Graph-Po ranks the 1$^{st}$ place in 3D AP and BEV AP with competitive inference speed.
Graph-VoI outperforms all previous multi-modality methods by a large margin (+2.08\% for 3D easy AP and +2.6\% for 3D moderate AP).
Table~\ref{kitti_val_r40} shows our method could consistently improve PointPillars, SECOND, and 3DSSD by a large margin, demonstrating the efficacy of the method.

\begin{table}[!t]
	\centering
	\caption{Performance comparison on the KITTI testing sever for 3D car detection. L and I represent the LiDAR point cloud and the camera image, respectively.}
	\resizebox{0.55\columnwidth}!{\begin{tabular}{c|c|ccc|ccc|c}
			\toprule
			\multirow{2}{*}{Methods} & \multirow{2}{*}{Modality} & \multicolumn{3}{c|}{3D AP} & \multicolumn{3}{c|}{BEV AP} & \multirowcell{2}{FPS\\(Hz)} \\
			& & Easy & Moderate & Hard & Easy & Moderate & Hard \\
			\midrule
			\bf{One-stage:} & & & & & & & \\
			% SECOND~\cite{yan2018second} & L & 83.34 & 72.55 & 65.82 & 89.39 & 83.77 & 78.59 & 20.0 \\
			% PointPillars~\cite{lang2019pointpillar} & L & 82.58 & 74.31 & 68.99 & 90.07 & 86.56 & 82.81 & \bf{42.4} \\
			Point-GNN~\cite{shi2020pointgnn} & L & 88.33 & 79.47 & 72.29 & 93.11 & 89.17 & 83.90 & 1.7 \\
			3DSSD~\cite{yang20203dssd} & L & 88.36 & 79.57 & 74.55 & 92.66 & 89.02 & 85.86 & 26.3 \\ 
			SA-SSD~\cite{he2020sassd} & L & 88.75 & 79.79 & 74.16 & 95.03 & 91.03 & 85.96 & 25.0 \\
			CIA-SSD~\cite{zheng2020ciassd} & L & 89.59 & 80.28 & 72.87 & 93.74 & 89.84 & 82.39 & \textbf{32.5} \\
			SASA~\cite{chen2022sasa} & L & 88.76 & 82.16 & 77.16 & 92.87 & 89.51 & 86.35 & 27.8 \\
			\midrule
			\bf{Two-stage:} & & & & & & & \\
			% PointRCNN~\cite{shi2019pointrcnn} & L & 86.96 & 75.64 & 70.70 & 92.13 & 87.39 & 82.72 & 10.0 \\
			% STD~\cite{yang2019std} & L & 87.95 & 79.71 & 75.09 & 94.74 & 89.19 & 86.42 & 12.5 \\
			PV-RCNN~\cite{shi2020pvrcnn} & L & 90.25 & 81.43 & 76.82 & 94.98 & 90.65 & 86.14 & 12.5 \\
			Voxel R-CNN~\cite{deng2021voxelrcnn} & L & 90.90 & 81.62 & 77.06 & 94.85 & 88.83 & 86.13 & 25.2 \\
			CT3D~\cite{sheng2021ct3d} & L & 87.83 & 81.77 & 77.16 & 92.36 & 88.83 & 84.07 & 14.3 \\
			Pyramid-PV~\cite{mao2021pyramid} & L & 88.39 & 82.08 & 77.49 & 92.19 & 88.84 & 86.21 & 7.9 \\
			VoTr-TSD~\cite{mao2021votr} & L & 89.90 & 82.09 & \textbf{79.14} & 94.03 & 90.34 & 86.14 & 7.2 \\
			SPG~\cite{xv2021spg} & L & 90.50 & 82.13 & 78.90 & 94.33 & 88.70 & 85.98 & 6.4 \\
			PointPainting~\cite{vora2020pointpainting} & L+I & 82.11 & 71.70 & 67.08 & 92.45 & 88.11 & 83.36 & 2.5 \\
			PI-RCNN~\cite{xie2020pircnn} & L+I & 84.37 & 74.82 & 70.03 & 91.44 & 85.81 & 81.00 & 10.0 \\
			MMF~\cite{liang2019mmf} & L+I & 88.40 & 77.43 & 70.22 & 93.67 & 88.21 & 81.99 & 12.5 \\
			EPNet~\cite{huang2020epnet} & L+I & 89.81 & 79.28 & 74.59 & 94.22 & 88.47 & 83.69 & 10.0 \\
			3D-CVF~\cite{yoo20203dcvf} & L+I & 89.20 & 80.05 & 73.11 & 93.52 & 89.56 & 82.45 & 13.3 \\
			CLOCs\_PVCas~\cite{pang2020clocs} & L+I & 88.94 & 80.67 & 77.15 & 93.05 & 89.80 & 86.57 & 10.0 \\
			\midrule
			\textbf{Graph-Pi (Ours)} & L & 90.94 & 82.42 & 77.00 & 95.06 & 91.52 & 86.42 & \textbf{28.5} \\ 
			\textbf{Graph-Vo (Ours)} & L & 91.29 & 82.77 & 77.20 & 95.27 & 91.72 & 86.51 & 25.6 \\ 
			\textbf{Graph-Po (Ours)} & L & 91.79 & 83.18 & 77.98 & \textbf{95.79} & \textbf{92.12} & \textbf{87.11} & 16.1 \\
			\textbf{Graph-VoI (Ours)} & L+I & \textbf{91.89} & \textbf{83.27} & 77.78 & 95.69 & 90.10 & 86.85 & 13.3 \\
			\bottomrule 
	\end{tabular}}
	\label{kitti_test}
\end{table}

\begin{table}[t]
	\centering
	\caption{Performance of our model on the KITTI $val$ set with AP calculated by 40 recall positions for car class. $^\dag$ indicates our reproduced results.}
	\resizebox{0.5\columnwidth}!{\begin{tabular}{c|ccc|ccc}
			\toprule 
			\multirow{2}{*}{Methods} & \multicolumn{3}{c|}{3D AP} & \multicolumn{3}{c}{BEV AP} \\
			& Easy & Moderate & Hard & Easy & Moderate & Hard \\
			\midrule
			Pointpillars$^\dag$ (\textbf{Pi}llar-based)~\cite{lang2019pointpillar} & 89.67 & 80.38 & 78.80 & 93.56 & 89.53 & 88.57 \\
			% 			\textbf{Graph-Pi (Ours)} & 93.17 & 85.73 & 82.94 & 96.19 & 91.90 & 89.29 \\
			\textbf{Graph-Pi (Ours)} & 93.16 & 85.87 & 83.29 & 96.18 & 91.84 & 89.46 \\
			\midrule
			SECOND$^\dag$ (\textbf{Vo}xel-based)~\cite{yan2018second} & 92.15 & 82.43 & 79.26 & 95.78 & 91.26 & 88.57 \\
			\textbf{Graph-Vo (Ours)} & 93.33 & 86.12 & 83.29 & 96.35 & 92.16 & 91.54  \\
			\textbf{Graph-VoI (Ours)} & \textbf{95.67} & \textbf{86.87} & \textbf{84.09} & 96.28 & \textbf{92.68} & \textbf{92.11}  \\
			\midrule
			3DSSD$^\dag$ (\textbf{Po}int-based)~\cite{yang20203dssd} & 91.68 & 82.72 & 79.74 & 96.04 & 91.45 & 88.89 \\
			\textbf{Graph-Po (Ours)} & 93.27 & 86.50 & 83.87 & \textbf{96.64} & 92.45 & 89.92 \\
			\bottomrule 
	\end{tabular}}
	\label{kitti_val_r40}
\end{table}

\begin{table}[t]
	\centering
	\caption{Ablation study of every module: RoI-graph pooling (RGP), multi-level attentive fusion (MLAF), dynamic point aggregation (DPA), PointNet (PN), and diagonal corners (DC). We show the 3D APH at level 2 on the WOD validation set.}
	\resizebox{0.5\columnwidth}!{\begin{tabular}{ccccc|cc}
			\toprule
			w/ RGP & w/ MLAF & w/ DPA & w/ PN & w/ DC & Vehicle & Pedestrian \\
			\midrule
			& &  &  & & 66.17 & 62.67 \\
			\checkmark & &  &  & & 69.48 & 66.79 \\
			\checkmark & \checkmark & & & & 69.77 & 67.00 \\
			\checkmark & \checkmark & \checkmark & & & 70.83 & 67.87 \\
			\checkmark & \checkmark & \checkmark & \checkmark & & 71.04 & 67.96 \\
			\checkmark & \checkmark & \checkmark & \checkmark & \checkmark & \textbf{72.10} & \textbf{69.02} \\
			\bottomrule
	\end{tabular}}
	\label{modules_ablation}
\end{table}

\begin{table*}[!t]
	\begin{minipage}[b]{0.48\linewidth}
	\centering
	\caption{Ablation study of the performance of DPA at different distances. We show the level 2 APH on the WOD validation set for vehicle class.}
	\resizebox{0.68\columnwidth}!{\begin{tabular}{c|cccc}
			\toprule
			w/ DPA & Overall & 0-30m & 30-50m & 50m-Inf \\
			\midrule
			& 69.77 & 90.25 & 71.21 & \textbf{45.40}   \\
			\checkmark & \textbf{70.83} & \textbf{91.74} & \textbf{71.36}  & \textbf{45.40}  \\
			\bottomrule
	\end{tabular}}
	\label{dpa_dis}
	\end{minipage}
	\hfill
	\begin{minipage}[b]{0.48\linewidth}
	\centering
	\caption{Ablation study of FPS and DFVS based on 3DSSD. We report the mAP on KITTI $val$ set for car class and the runtime of sampling.}
    \resizebox{0.70\columnwidth}!{\begin{tabular}{c|ccc}
			\toprule
			Methods & 3D mAP & BEV mAP & Runtime (ms) \\
			\midrule
			FPS & 81.73 & \textbf{88.57} & 29.6 \\ 
			DFVS & \textbf{81.75} & 88.55 & \textbf{20.3} \\ 
			\bottomrule
	\end{tabular}}
	\label{dfvs_in_3dssd}
	\end{minipage}
\end{table*}

\begin{table}[!t]
	\centering
	\caption{Ablation study of different sampling and searching strategies. $^\dag$ is tested by us based on the officially released code.}
	\resizebox{0.55\columnwidth}!{\begin{tabular}{cccccc|cc}
			\toprule
			w/ PR$^\dag$ & w/ PS & w/ VS & w/ DVS & w/ DFVS & w/ FPS & APH & Runtime (ms) \\
			\midrule
			\checkmark & & & & & & 69.77 & 69.7 \\
			& \checkmark & & & & & 69.77 & \textbf{1.9} \\ 
			& \checkmark & \checkmark & & & & 70.39 & 2.2  \\
			& \checkmark & & \checkmark & & & 70.67 & 2.2 \\
			& \checkmark & & & \checkmark & & \textbf{70.83} & 2.8 \\
			& \checkmark & & & & \checkmark & \textbf{70.83} & 7.3 \\
			\bottomrule
	\end{tabular}}
	\label{sampling_abla}
\end{table}

\subsection{Ablation Study} \label{subsection:as}
\paragraph{Analysis of the Dynamic Point Aggregation.}
The third and fourth rows in Table~\ref{modules_ablation} show that the dynamic point aggregation (DPA) contributes an improvement of 1.06\% and 0.87\% APH at level 2 for vehicle and pedestrian, respectively.
Especially, Table~\ref{dpa_dis} shows that DPA improves the baseline by 1.49\% APH at 0-30m since the nearby objects suffer more from the uneven distribution problem.
The first and second rows in Table~\ref{sampling_abla} show that the patch search (PS) is 35$\times$ faster than the baseline, i.e., point cloud region pooling (PR)~\cite{shi2019pointrcnn}.

We also explore several sampling strategies in Table~\ref{sampling_abla} to solve the uneven distribution problem, i.e., voxel sampling (VS), dynamic voxel sampling (DVS), dynamic farthest voxel sampling (DFVS), and farthest point sampling (FPS).
We note that VS is a special case of DVS when $\delta$ is large, and FPS is a special case of DFVS when $\lambda$ is small. Table~\ref{sampling_abla} shows that DFVS achieves the best trade-off between accuracy and efficiency. Further, we explore the use of DFVS on point-based 3D object detectors as an alternative to FPS. Table~\ref{dfvs_in_3dssd} shows that DFVS achieves similar results with FPS but costs less runtime.

\paragraph{Analysis of the RoI-graph Pooling.}
The first and second rows in Table~\ref{modules_ablation} show that the RoI-graph pooling (RGP) raises the APH at level 2 for vehicle and pedestrian by 3.31\% and 4.12\%, respectively.
In Table~\ref{gnn_setup}, we study the effect of the number of iterations on the detection accuracy, where the number of neighbors is set to $8$ by default to save GPU memory.
This result suggests it is beneficial to iterate more times to mine geometric relations.
Besides, we find accuracy gains for distant objects are greater than for nearby objects since the contextual information is better modeled to alleviate the missing detection of distant objects.
For graph construction, we investigate the components used in the initial state of the graph node.
The fourth and fifth rows in Table~\ref{modules_ablation} show that using PointNet (PN) raises 0.21\% APH for vehicle class since the downsampling introduces the loss of fine-grained details, and the fifth and sixth rows show that adding two diagonal corners (DC) for each node raises 1.06\% APH for both vehicle and pedestrian.
For graph aggregation, the second and third rows in Table~\ref{modules_ablation} show that multi-level attentive fusion (MLAF) boosts the performance by 0.29\% APH for vehicle class.
Furthermore, we study influences of different aggregation methods in Table~\ref{agg}, i.e., channel-wise Transformer (CT), Set Transformer (ST), attention sum (AS), average pooling (AP), and max pooling (MP).
Transformer does not achieve better results, probably because it has more parameters leading to overfitting.

\begin{table*}[!t]
	\begin{minipage}[b]{0.48\linewidth}
	\centering
	\caption{Ablation study of the number of iterations to update the graph.}
	\resizebox{0.70\columnwidth}!{\begin{tabular}{c|cccc}
	\toprule
	\# iterations & Overall & 0-30m & 30-50m & 50m-Inf \\
	\midrule
	T=1 & 68.76 & 89.84 & 70.19 & 44.03 \\ 
	T=2 & 69.07 & 89.95 & 70.60 & 44.49 \\ 
	T=3 & \textbf{69.48} & \textbf{89.97} & \textbf{71.02} & \textbf{45.18} \\
	\bottomrule
	\end{tabular}}
	\label{gnn_setup}
	\end{minipage}
	\hfill
	\begin{minipage}[b]{0.48\linewidth}
	\centering
	\caption{Ablation study of different methods to aggregate nodes' features.}
	\resizebox{0.68\columnwidth}!{\begin{tabular}{c|ccccc}
			\toprule
			Methods & CT & ST & AS & AP & MP \\
			\midrule
			Vehicle & 71.67 & 71.82 & 71.86 & 71.85 & \textbf{72.10}\\ 
			Pedestrian & 68.82 & 68.74 & 68.66 & 68.76 & \textbf{69.02}\\
			\bottomrule
	\end{tabular}}
	\label{agg}
	\end{minipage}
\end{table*}

\begin{table}[!t]
	\centering
	\caption{Ablation study of image features and CPA. $^\dag$ and $^\ddag$ indicate CPA is used in RPN and refinement, respectively.}
	\resizebox{0.48\columnwidth}!{\begin{tabular}{ccccc|cc}
	\toprule
	w/ CPA$^\dag$ & w/ CPA$^\ddag$ & w/ RGB & w/ Seg. & w/ Feat. & 3D AP & BEV AP \\
	\midrule
	& & & & & 85.38 & 91.48 \\
	\checkmark & \checkmark & & & & 86.12 & 92.16 \\
	\checkmark & & & & & 86.11 & 92.23 \\
	\checkmark & & \checkmark & & & 86.20 & 92.35 \\
	\checkmark & & & \checkmark & & 86.38 & 92.59 \\
	\checkmark & & & & \checkmark & \textbf{86.87} & \textbf{92.68}\\
	\bottomrule
	\end{tabular}}
	\label{img_stream}
\end{table}

\paragraph{Analysis of the Visual Features Augmentation.}
By comparing the fourth and fifth rows in Table~\ref{kitti_val_r40}, we observe that the image feature raises the detection results by 2.34\%, 0.75\%, and 0.8\% 3D AP respectively in terms of easy, moderate, and hard.
We carefully analyze the effect of the cut-and-paste augmentation (CPA) at different stages in Table~\ref{img_stream}.
We find that the refinement stage is hardly affected by CPA.
Thus, we can conveniently train the LiDAR branch and the image branch end-to-end without the help of CPA.
We also provide the study of different image features, i.e., the RGB of the input image, the segmentation scores, and the output features of the 2D detector.
We find that using the 2D features achieves the best results.

\section{Conclusions}
We present an accurate and efficient 3D object detector Graph R-CNN that can be applied to existing 3D detectors.
Our framework can handle the unevenly distributed and sparse point clouds by utilizing the dynamic point aggregation and the semantic-decorated local graph.
% In the future, we plan to explore the point cloud completion to compensate for the sparse points to capture robust RoI features and knowledge distillation to improve the detection performance.

\paragraph{Acknowledgments.} This work was supported in part by The National Key Research and Development Program of China (Grant Nos: 2018AAA0101400), in part by The National Nature Science Foundation of China (Grant Nos: 62036009, U1909203, 61936006, 62133013), in part by Innovation Capability Support Program of Shaanxi (Program No. 2021TD-05).

\clearpage
\bibliographystyle{splncs04}
\bibliography{egbib}
\end{document}